%% file: SinkhornDivergences.tex
\documentclass[twoside]{article}
\usepackage{aistats2018}
\pdfoutput=1

\newcommand{\final}[2]{#1}

\usepackage{aistats2018}

\usepackage{caption,subcaption}
\usepackage{wrapfig}
\usepackage[utf8]{inputenc} 
\usepackage[T1]{fontenc}    
\usepackage{hyperref}       
\usepackage{url}            
\usepackage{booktabs}       
\usepackage{amsfonts}       
\usepackage{nicefrac}       
\usepackage{microtype}      
\usepackage{mystyle,amssymb,dsfont,amsmath,amsthm,mathtools}

\graphicspath{{./images/}}

\begin{document}

\twocolumn[

\aistatstitle{Learning Generative Models with Sinkhorn Divergences}

\final{
\aistatsauthor{ Aude Genevay \And Gabriel Peyr\'e \And  Marco Cuturi }
\aistatsaddress{ CEREMADE, \\ Universit\'e Paris-Dauphine \And  CNRS and DMA, \\ \'Ecole Normale Sup\'erieure \And ENSAE CREST \\    Universit\'e Paris-Saclay }
}{
\aistatsauthor{ Author 1 \And Author 2 \And  Author 3 }
\aistatsaddress{ Institution 1 \And  Institution 2 \And Institution 3 }
}

]

\input{sections/abstract}
\input{sections/intro}
\input{sections/density-fitting}
\input{sections/sinkhorn}

\input{sections/applications}

\input{sections/conclusion}

\bibliographystyle{plain}
\bibliography{biblio}
\normalsize


\final{
\appendix
\input{sections/numerics-div}

}{}

\end{document}

%% file: sections/abstract.tex

\begin{abstract}
The ability to compare two degenerate probability distributions, that is two distributions supported on low-dimensional manifolds in much higher-dimensional spaces, is a crucial factor in the estimation of generative models.
It is therefore no surprise that optimal transport (OT) metrics and their ability to handle measures with non-overlapping supports have emerged as a promising tool. Yet, training generative machines using OT raises formidable computational and statistical challenges, because of \emph{(i)} the computational burden of evaluating OT losses, \emph{(ii)} their instability and lack of smoothness, \emph{(iii)} the difficulty to estimate them, as well as their gradients, in high dimension.
This paper presents the first tractable method to train large scale generative models using an OT-based loss called Sinkhorn loss which tackles these three issues by relying on two key ideas: \emph{(a)} entropic smoothing, which turns the original OT loss into a differentiable and more robust quantity that can be computed using Sinkhorn fixed point iterations; \emph{(b)} algorithmic (automatic) differentiation of these iterations with seamless GPU execution.
%
%
Additionally, Entropic smoothing generates a family of losses interpolating between Wasserstein (OT) and Energy distance/Maximum Mean Discrepancy (MMD) losses, thus allowing to find a sweet spot leveraging the geometry of OT on the one hand, and the favorable high-dimensional sample complexity of MMD, which comes with unbiased gradient estimates.
The resulting computational architecture complements nicely standard deep network generative models by a stack of extra layers implementing the loss function. 
\end{abstract}

%% file: sections/intro.tex

\section{Introduction}

Several important statistical problems boil down to fitting densities, \emph{i.e.} estimating the parameters of a chosen model that \emph{fits} observed data in some meaningful way.
While the standard approach is maximum likelihood estimation, this approach is often flawed in machine learning tasks where the sought after distribution is obtained in a generative fashion, \emph{i.e.} described using a sampling mechanism (often a non-linear function mapping a low dimensional latent random vector to a high dimensional space). Indeed, in these settings, the density is singular in the sense that it only has positive probability on a low-dimensional manifold of the observation space and is zero elsewhere. 
%
To remedy these issues, and in line with several recent proposals~\cite{bassetti2006minimum,montavon2016wasserstein,bernton2017inference,WassersteinGAN}, we propose to shift away from information divergence based methods (among which the MLE) and consider instead the geometry of optimal transport~\cite{villani2003,santambrogio2015optimal} to define such a fitting criterion.

\begin{figure*}[ht]
\centering
	\includegraphics[width=\linewidth]{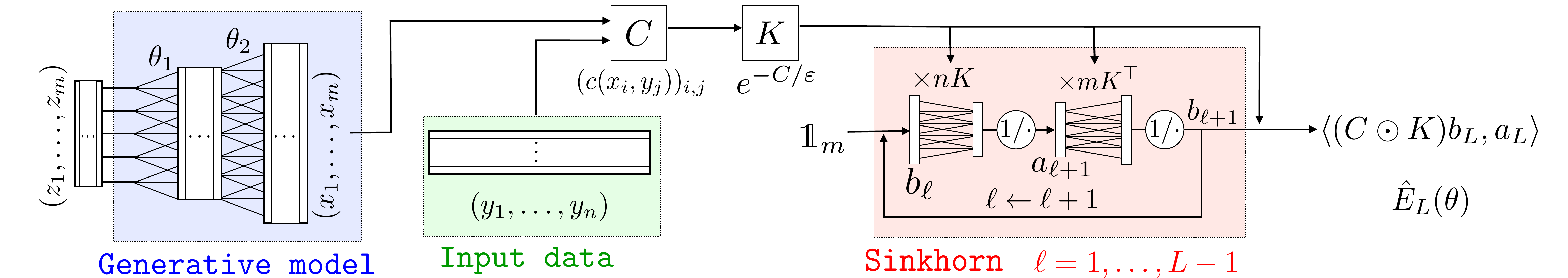}
\caption{\label{fig-workflow} %
	For a given fixed set of samples $(z_1,\ldots,z_m)$, and input data $(y_1,\ldots,y_n)$, flow diagram for the computation of Sinkhorn loss function $\th \mapsto \hat E_\epsilon^{(L)}(\th)$. This function is the one on which automatic differentiation is applied to perform parameter learning. The display shows a simple 2-layer neural network $g_\th : z \mapsto x$, but this applies to any generative model.
	}
\end{figure*}

\paragraph{Previous works.}

For purely generative models, several likelihood-free workarounds exist. 
Major approaches include variational autoencoders (VAE)~\cite{VAE}, generative adversarial networks (GAN)~\cite{GAN} and several more variations including combinations of both~\cite{pmlr-v48-larsen16}. The adversarial GAN approach is implicitly geometric in the sense that it computes the best achievable classification accuracy (taking for granted the training and generated datapoints have opposite labels) for a given class of classifiers as a proxy for the distance between two distributions: If accuracy is high distributions are well separated, if accuracy is low they are difficult to tell apart and lie thus at a very close distance.

Geometry was also explicitly considered when trying to minimize a flexible metric between distributions: the maximal mean discrepancy~\cite{gretton2007kernel}. It was shown in ensuing works that the effectiveness of the MMD in that setting~\cite{li2015generative,MMD-GAN} hinges on the ability to find a relevant RKHS bandwidth parameter, which is a highly nontrivial choice.
The Wasserstein or earth mover's distance, long known to be a powerful tool to compare probability distributions with non-overlapping supports, has recently emerged as a serious contender to train generative models.
While it was long disregarded because of its computational burden---in its original form solving OT amounts to solving an expensive network flow problem when comparing discrete measures in metric spaces---recent works have shown that this cost can be largely mitigated by settling for cheaper approximations obtained through strongly convex regularizers, in particular entropy~\cite{CuturiSinkhorn,2016-genevay-nips}. The benefits of this regularization has opened the path to many applications of the Wasserstein distance in relevant learning problems~\cite{courty2014domain,2015-Frogner,NIPS2016_6139,pmlr-v51-rolet16}. 
Although the use of Wasserstein metrics for inference in generative models was considered over ten years ago in~\cite{bassetti2006minimum}, that development remained exclusively theoretical until a recent wave of papers managed to implement that idea more or less faithfully using several workarounds: entropic regularization over a discrete space~\cite{montavon2016wasserstein}, approximate Bayesian computations~\cite{bernton2017inference} and a neural network parameterization of the dual potential arising from the dual OT problem when considering the 1-Wasserstein distance~\cite{WassersteinGAN}.
As opposed to this dual way to compute gradients of the fitting energy, we advocate for the use of a primal formulation, which is numerically stable, because it does not involve differentiating the (dual) solution of an OT sub-problem, as also pointed out in~\cite{Bousquet2017}.
Additionally, introducing entropic regularization in the formulation of optimal transport allows to interpolate between a pure OT loss and a Maximum Mean Discrepency loss, thus bridging the gap between these two approaches often presented as opposed points of view.

\paragraph{Contributions.}

The main contributions of this paper are twofold : (i) a theoretical comtribution regarding a new OT-based loss for generative models, (ii) a simple numerical scheme to learn under this loss. (i) We introduce the Sinkhorn loss, based on regularized optimal transport with an entropy penalty, and we prove that when the smoothing parameter $\epsilon=+0$ we recover pure OT loss whereas letting $\epsilon=+\infty$ leads to MMD. The addition of entropy is important to reduce sample complexity and gradient bias, and thus allows us to take advantage of the good geometrical properties of OT without its drawbacks in high-dimensions. (ii) We propose a computationally tractable and stable approach to learn with that Sinkhorn loss, which enables inference for any differentiable generative model. It operates by adding $L$ additional pooling layers (application of a filtering kernel $K$ and pointwise divisive non-linearities), as illustrated on Figure~\eqref{fig-workflow}. As routinely done in standard deep-learning architecture frameworks, the training is then achieved using stochastic gradient descent and automatic differentiation. This provides accurate and stable approximation of the loss and its gradient, at a reasonable extra computational cost, and streams nicely on GPU hardware. 

\paragraph{Notations.}

For a matrix $A$, $A^\top$ denotes its transpose. 
For two vectors (or matrices) $\dotp{u}{v} \eqdef \sum_i u_i v_i$ is the canonical inner product (the Frobenius dot-product for matrices).
We define $\ones_{\m} \eqdef (1/\m,\ldots,1/\m) \in \RR_+^{\m}$ the uniform histogram, so that for $\P \in \RR^{\n \times \m}$, $\P \ones_{\m} \in \RR^\n$ and $\P^\top \ones_{\n} \in \RR^\m$ stand for the row and column averages of $P$.
We denote $\Mm_+^1(\Xx)$ the set of probability distributions (positive Radon measures of unit mass) over a metric space $\Xx$. $\de_x$ stands for the Dirac (unit mass) distribution at point $x \in \Xx$.
For some continuous map $g : \Zz \rightarrow \Xx$, we denote $g_\sharp : \Mm_+^1(\Zz) \rightarrow \Mm_+^1(\Xx)$ the associated push-forward operator, which is a linear map between distributions. This corresponds to defining, for $\zeta \in \Mm_+^1(\Zz)$ and $B \subset \Xx$, $(g_\sharp \zeta)(B) = g^{-1}(B)$ ; or equivalently, that $\int_\Xx \phi \d(g_\sharp \zeta) = \int_\Zz \phi \circ g \d \zeta$ for continuous functions $\phi$ on $\Xx$ ; or equivalently that a random sample $x$ from $g_\sharp \zeta$ can be obtained as $x=g(z)$ where $z$ is a random sample from $\zeta$. 

%% file: sections/density-fitting.tex

\section{Minimum Kantorovich Estimation}

\paragraph{Density fitting.}

We consider a data set of $\N$ (usually very large) observations $(y_1,\dots,y_\N) \in \Xx^\N$ and we want to learn a generative model that produces samples that are similar to that dataset. Samples $x=\gt(z)$ from the generative model are defined by taking as input a sample $z \in \Zz$ from some reference measure $\muz$ (typically a uniform or a Gaussian measure in a low-dimensional space $\Zz$) and mapping it through a differentiable function $\gt : \Zz \rightarrow \Xx$. Formally, this corresponds to defining the generative model measure $\mut$ from which $x$ is drawn as $\mut = g_{\th\#}\muz$. Our goal is to find $\theta$ which minimizes a certain loss $\Loss$ between $\mut$ and the empirical measure $\nu$ associated with the data
\eql{\label{eq-fitting-generic}
	\theta \in \uargmin{\theta} \Loss(\mut,\nu)  
	\qwhereq 
	\nu \eqdef \frac{1}{\N} \sum_{j=1}^{\N} \delta_{y_j}.
}
While we focus here for simplicity on the case of deterministic encoding functions $g_\th$ between $\muz$ and $\mut$, our method extends to more general probabilistic generative models, such as VAE~\cite{VAE}.

\paragraph{Distances between measures.}
Maximum likelihood estimation (MLE) is obtained by setting $\Loss(\mut,\nu) = -\sum_j \log \frac{\d\mut}{\d x}(y_j)$, where $\frac{\d\mu}{\d x}$ is the density of $\mut$ with respect to a fixed reference measure (a typical choice is $\d x$ being the Lebesgue measure in $\Xx=\RR^d$). This MLE loss can be seen as a discretized version of the relative entropy (a.k.a. the Kullback-Leibler divergence). A major issue with this approach is that in general generative models defined this way (when $\Zz$ has a much smaller dimensionality than $\Xx$) have singular distributions (\emph{i.e.} supported on a low-dimensional manifold), without density with respect to a fixed measure, and therefore MLE cannot be considered.

The usual workaround is to assume that $\Xx$ is equipped with some distance $d_\Xx$, and consider weak metrics, which take into account spatial displacement of these measures, enabling the comparison of singular measures. A classical construction for such a loss function $\Loss$ is through duality (see \emph{e.g}~\cite{sriperumbudur2012empirical}), namely by considering a dual norm $\Loss(\mu,\nu)=\norm{\mu-\nu}_B^*$ where $\norm{\xi}_B^* = \sup\enscond{ \int_\Xx h(x)\d\xi(x) }{ h \in B }$. Here $B$ is a ``unit ball'' of continuous functions that should contain $0$ in its interior. This ensures that $\norm{\cdot}_B^*$ is well defined even for singular inputs, and it is a norm which metrizes the weak convergence of measures (so that for instance $\Loss(\de_x,\de_{x'}) \rightarrow 0$ as $x \rightarrow x'$), see~\cite[Sec.7.2.1]{santambrogio2015optimal} for more details. Classical instances of such settings include the 1-Wasserstein distance (obtained by setting $B = \enscond{g}{\norm{\nabla g}_\infty \leq 1}$ the set of 1-Lipschitz functions) and reproducing kernel Hilbert spaces (letting $B=\enscond{g}{\norm{k \star g}_{L^2(\Xx)} \leq 1}$ where $k$ is an appropriate convolution kernel). The latter define the class of Maximum Mean Discrepency losses \cite{gretton2007kernel} defined by
\begin{equation}
\begin{split}
	\norm{\mu,\nu}_k &= \mathbb{E}_{\mu\otimes\mu}[k(X,X')] + \mathbb{E}_{\nu\otimes \nu}[k(Y,Y')] \\
	&- 2 \mathbb{E}_{\mu\otimes \nu}[k(X,Y)]
\end{split}
\end{equation}  


\paragraph{Optimal transport distances.}

In this article, we advocate for a different approach, which is to consider generic optimal transport (OT) metrics which can be used over general spaces $\Xx$ (not just the Euclidean space $\RR^d$ and not only the 1-Wasserstein distance). 
The OT metric between two probability distributions $(\mu,\nu) \in \Mm_+^1(\Xx) \times \Mm_+^1(\Xx)$ supported on two metric spaces $(\Xx,\Xx)$ is defined as the solution of the (possibly infinite dimensional) linear program:
\eql{\label{primal}
	\Ww_c(\mu,\nu) \eqdef \min_{\pi \in\Pi(\mu,\nu) } \int_{\Xx \times \Xx} c(x,y) \d \pi(x,y),
}
where the set of couplings is composed of joint probability distributions over the product space $\Xx \times \Xx$ with imposed marginals $(\mu,\nu)$
\eq{ 
	\Pi(\mu,\nu) \eqdef \enscond{\pi \in \Mm_+^1(\Xx \times \Xx) }{ P_{1\sharp}\pi=\mu, P_{2\sharp}\pi=\nu  },
} 
where $P_1(x,y)=x, P_2(x,y)=y$ are simple projector operators.
Formula~\eqref{primal} corresponds to the celebrated Kantorovitch formulation~\cite{Kantorovich42} of OT (see~\cite{santambrogio2015optimal} for a detailed account on the theory). 
Here $c(x,y)$ is the ``ground cost'' to move a unit of mass from $x$ to $y$, and we shall make no assumptions (except for regularity) on its form. When $\Xx$ is equipped with a distance $d_\Xx$, a typical choice is to set $c(x,y)=d_\Xx(x,y)^p$ where $p>0$ is some exponent, in which case for $p \geq 1$ $\Ww_c^{1/p}$ is the so-called $p$-Wasserstein distance between probability measures.

We introduce the regularized optimal transport problem~\cite{CuturiSinkhorn,2016-genevay-nips} defined by
\begin{equation}\label{OTreg}
\min_{\pi\in\Pi(\mu,\nu)} \int c(x,y) \d\pi(x,y) + \epsilon \int \log(\frac{\pi(x,y)}{\d\mu(x)\d\nu(y)})\d\pi(x,y) \tag{$\Pp_\epsilon$}
\end{equation}

And the associated regularized Wasserstein distance associated with cost $c$ and regularization paremeter $\epsilon$ is defined by:
$$\Ww_{c,\epsilon} (\mu,\nu) = \int c(x,y) \d \pi_\epsilon(x,y)$$ where $\pi_\epsilon$ is the optimal coupling for the regularized OT problem \eqref{OTreg}.
 
\begin{thm}[Sinkhorn Loss]
The Sinkhorn loss between two measure $\mu,\nu$ is defined as: \eql{\label{eq-sinkh-loss}\bar{\Ww}_{c,\epsilon} (\mu,\nu) = 2 \Wce (\mu,\nu) - \Wce (\mu,\mu) -\Wce (\nu,\nu).} 
with the following limiting behavior in $\epsilon$: \begin{enumerate}
 \item as $\epsilon \rightarrow 0$,\quad $ \bar{\Ww}_{c,\epsilon} (\mu,\nu) \rightarrow 2 \Ww_c(\mu,\nu)  $
 \item as $\epsilon \rightarrow +\infty$,\quad $ \bar{\Ww}_{c,\epsilon} (\mu,\nu) \rightarrow MMD_{-c}(\mu,\nu) $
\end{enumerate}
where $MMD_{-c}$ is the MMD distance whose kernel is the cost from the optimal transport problem.
\end{thm}

\begin{rem}This theorem is a generalization of~\cite[\S3.3]{e19020047} for continuous measures. \end{rem}

\begin{proof}
1. The first part of the assumption is well known, see for instance \cite{Carlier2017}.

2. Letting $\epsilon$ go to infinity in the regularized OT problem amounts to finding the coupling with minimum entropy in the constraint set. The problem becomes $\min_{\pi \in \Pi(\mu,\nu)} \int log(\frac{\pi(x,y)}{\d\mu(x)\d\nu(y)})d\pi(x,y)$ where $\Pi(\mu,\nu)$ is the set of couplings with marginals $\mu$ and $\nu$. Introducing Lagrange multipliers $u$ and $v$ for these constraints, the dual problem becomes $\max_{u,v} \int u(x) \d\mu(x) + \int v(y) \d\nu(y) - \int \exp(u(x)+v(y))\d\mu(x) \d\nu(y)$ and the primal-dual relation is given by $d\pi(x,y) = \exp(u(x)+v(y))\d\mu(x) \d\nu(y)$. Solving the dual gives $u = v = 0$ and thus the optimal coupling is simply the product of the marginals i.e. $\pi = \mu \otimes \nu$.
\end{proof}

The density fitting problem can be rewritten using the Sinkhorn divergence~\eqref{eq-sinkh-loss}:
\eq{
\umin{\theta} E_\epsilon(\theta) \qwhereq  E_\epsilon(\theta) \eqdef \bar{\Ww}_{c,\epsilon} (\mut,\nu).
}

\paragraph{A Discussion on OT \emph{vs.} MMD}
As proved in Theroem 1, the Sinkhorn loss interpolates between a pure OT loss for $\epsilon=0$ and MMD losses for $\epsilon=+\infty$. 
As such, when $\epsilon \rightarrow +\infty$, our loss takes advantage of the good properties of MMD losses, and in particular a favorable sample complexity of $O(1/\sqrt{n})$ (decay rate of the approximation of the true loss with a mini-batch of size $n$) and unbiased gradient estimates when using mini-batches. Note that sample complexity estimates have not been proved for the Sinkhorn loss, but empirical evidence (see curves in supplementary material) shows that its behavior is similar to that of MMD when epsilon is not too small.  In contrast, the unregularized OT loss suffers from a sample complexity of $O(1/n^{1/d})$, see~\cite{weed2017sharp} for a recent account on this point. 
Using MMD to train generative models has been shown to be successful in~\cite{MMD-GAN,li2015generative}. The improved Wasserstein GAN approach~\cite{gulrajani2017improved} (which penalizes the squared norm of the gradient of the dual potential) is similar to an MMD (in fact a dual Sobolev norm).
By tuning the $\epsilon$ parameter, our method is able to take the best of both worlds, to blend the non-flat geometry of OT with the high-dimensional rigidity of MMD losses. Additionally, the Sinkhorn loss, as is the case for the original OT problem, can be defined with any cost $c$, whereas MMD losses are only meaningful when used with positive definite kernels $k$. The postivity of the Sinkhorn loss is yet to be proved but empirical evidence (see supplementary) strongly points in that direction.
Eventually, in the specific case where $c = \norm{\cdot}_p$ for $1<p<2$, the associated MMD loss is the energy distance~\cite{szekely2004testing}. It was also used to fit generative models in~\cite{CramerGAN}, while \cite{MMDGAN} uses MMD with a gaussian kernel. Note that contrary to what~\cite{CramerGAN} claims, the energy distance cannot be presented as a cure to solve the bias of OT estimation in high-dimension, since the two distances are fundamentally different. 

%% file: sections/sinkhorn.tex

\section{Sinkhorn AutoDiff Algorithm}

Computing an approximation of $\nabla E$ is itself a difficult problem, even when $\epsilon=0$. 
In the latter case, a workaround is to use, instead of differentiating the ``primal'' formula~\eqref{primal}, the optimum of the ``dual'' formula, resulting in
$\nabla E_0(\th) = \int_\Zz \nabla[ h \circ g_\th ](z) \d\muz(z)$, where $h$ is an optimal dual continuous potential for $\mu=\mut$, see~\cite{WassersteinGAN}.  
This requires the use of approximate semi-discrete OT solvers (because $\mut$ is a continuous measure while $\nu$ is discrete), which typically operate by approximating the continuous dual potential $h$, see for instance~\cite{2016-genevay-nips} which uses an RKHS expansion, or~\cite{WassersteinGAN} which uses a deep-network expansion.
While the dual formalism is appealing (in particular because it involves only integration over $\Zz$ and not the product space $\Zz \times \Xx$), the resulting gradient formula requires differentiating the dual potential, which tends to be difficult to compute and unstable. A very similar conclusion is reached by~\cite{Bousquet2017} (see in particular their Proposition 3).

We propose a different route, by making two key simplifications: (i) approximate the function $E_\epsilon(\th)$ by a size-$(\m,\n)$ mini-batch sampling $\hat E_\epsilon(\th)$ to make it amenable to stochastic gradient descent ; (ii) approximate $\hat E_\epsilon(\th)$ by $L$-steps of the Sinkhorn algorithm~\cite{CuturiSinkhorn} to obtain an algorithmic loss $\hat E_\epsilon ^{(L)}(\th)$ which is amenable to automatic differentiation.

\paragraph{(i) Mini-batch sampling loss.}

We replace the initial functional $E_\epsilon(\th)$ by an expectation over mini-batches of size $(\m,\n)$, with leads to consider 
\eql{\label{eq-minibatches}
	 \umin{\th}
	\EE(\hat E_\epsilon(\th))\qwhereq E_\epsilon(\th)\eqdef  \Wce(\hat \mu_\theta,\hat\nu)}
	
	\eq{\qandq \choice{
		\hat \mu_\theta \eqdef \frac{1}{\m}\sum_{i=1}^{\m} \de_{x_i}, \\
		\hat \nu \eqdef \frac{1}{\n}\sum_{j \in J} \de_{y_j}, 
	}
	\:
	\choice{
		(z_i)_{i=1}^{\m} \overset{\text{i.i.d}}{\sim} \muz,\\
		\foralls i, \: x_i \eqdef g_\th(z_i), 
	}
}
The expectation is taken over the samples $(z_i)_{i=1}^\m$ (drawn independently according to $\muz$) and the indexes $J \subset \{1,\ldots,\N\}$ with $|J|=\n$. As $(\m,\n)$ increases, $\EE(\hat E_\epsilon)$ approaches $E_\epsilon$, and convergence of minimizers is studied in~\cite{bernton2017inference}.

At a given iterate of this stochastic gradient descent scheme (see pseudo-code~\ref{algo-sgd}), one draws a mini-batch $(z_i)_{i=1}^{\m} \overset{\text{i.i.d}}{\sim} \muz$ and a subset $J$ of observations, and aims at computing the gradient of 
\eql{\label{eq-linprog-finite}
	\hat E(\th) = 
		 \umin{\P \in \RR_+^{\m \times \n}} \enscond{
		 	 	\dotp{\P}{\tc}
			}{ \P \ones_{\m} = \ones_\n, \P^\top \ones_{\m} = \ones_\n }, 
}
where we defined $\tc \eqdef \begin{bmatrix}c(g_\th(z_i), y_j)\end{bmatrix}_{i,j}\in\RR^{\m\times
\n}$ (which depends on $\th$ because the $x_i$'s do).
Note that this is simply a rephrasing of~\eqref{primal} in the case where both input measures are discrete (sums of Dirac masses), so that couplings $\pi$ can be treated as matrices $\P \in \RR^{\m \times \n}$, namely $\pi=\sum_{i,j} \P_{i,j} \de_{(z_i,y_j)} \in \Mm_+^1(\Zz \times \Xx)$. 
 
\paragraph{(ii) Sinkhorn iterates.}

One major advantage of regularizing the optimal transport problem is that it becomes solvable efficiently using Sinkhorn's algorithm~\cite{Sinkhorn64} (when dealing with discrete measures), and leads to a differentiable loss function (as first noticed in~\cite{CuturiSinkhorn,CuturiDoucet}). Such a regularization is known to be equivalent to restricting the search space in~\eqref{eq-linprog-finite} to couplings having the so-called scaling form
\eq{
	\P_{i,j} = a_i K_{i,j} b_j
	\qwhereq
	K_{i,j} \eqdef  e^{-\tc_{i,j}/\epsilon}.}
Matrix $K$ is the so-called Gibbs kernel. Note that $K$ depends implicitly on $\th$ (because matrix $\tc$ does), and contains therefore all of the geometric information related to the ability of $\th$ to sample points near the dataset.
Starting with $b^{(0)} \eqdef \ones_{\m}$, $\ell \leftarrow 0$, Sinkhorn iterates read
\eql{\label{eq-sinkhorn}
\itt{a} \eqdef \frac{\ones_\n}{K \it{b}} 
\qandq
\itt{b} \eqdef \frac{\ones_\m}{K^\top \itt{a}} 
}
where $\frac{\cdot}{\cdot}$ denotes component-wise division. 
The main computational burden of~\eqref{eq-sinkhorn} are the matrix-vector multiplication, which stream extremely well on GPU architectures, and therefore nicely add to a typical deep network architecture with $L$ additional layer of linear operations ($K$ can be interpreted as a localized linear filtering) and entry-wise non-linear operations (here divisions). 

For a given budget $L$ of iterations, our final loss is then obtained by using $\IT{\P} \eqdef \diag(\IT{a})K\diag(\IT{b})$ as a proxy for the optimal transport coupling, and thus
\eql{\label{eq-return-value}
	\hat E_\epsilon ^{(L)}(\th) \eqdef \dotp{\tc}{\IT{\P}} = 
	\sum_{i=1}^{\m} \sum_{j=1}^\n \tc_{i,j} \ITi{a} \ITj{b} K_{i,j}}
where it is once again important to remind that $K,\tc,\IT{b},\IT{a}$ depend on $\th$. As $\epsilon \rightarrow 0$ and $L \rightarrow +\infty$, one can show that the $\IT{\P}$ computed by Sinkhorn's iterates approaches a solution to~\eqref{eq-linprog-finite}, with linear convergence rate (deteriorating as $\epsilon \rightarrow 0$), so that $\hat E_\epsilon ^{(L)}(\th)$ is a smooth proxy for $E_\epsilon(\th)$ which can be differentiated in a fast and stable way, while being accurate as $\epsilon\rightarrow 0$ and $(\m,\n,L)$ increase. 
It is however important to realize that for large scale and high dimensional learning applications, empirical considerations~\cite{CuturiSinkhorn,kusner2015word,2015-Frogner} suggest that, unlike relevant applications of the same scheme in graphics~\cite{solomon2015convolutional}, a relatively strong regularization---a large $\epsilon$---leads not only to more stable results but also faster convergence, so that the value for $L$ can be set quite low.

\paragraph{Learning the cost function}

Aside from the regularization parameter, a key element of the Sinkhorn loss is the choice of the ground cost $c$ on the data space. In some cases, using a simple metric such as the $L^2$ norm is sufficient to compare two data points, but when dealing with high-dimensional objects, choosing $c$ is more critical. In such cases, we  propose to learn the cost $c$ with the following parametrization
\eq{
	c_\phi(x,y) \eqdef \norm{f_\phi(x)-f_\phi(y)} \qwhereq f_\phi : \Xx \rightarrow \RR^p,
}
where $f_\phi$ can be modelled by a neural network, and can be seen as a feature extractor that reduces the dimensionality of $\Xx$ through a mapping onto $\RR^p$. 

Learning the cost function here is very similar to learning a parametric kernel in an MMD model, as done in \cite{MMDGAN}. The optimization problem becomes a min-max problem over $(\theta,\phi)$ instead of a simple minimization problem over $\theta$ 
\eq{
\min_\theta \max_\phi \bar{\Ww}_{c_\phi,\epsilon(\mut,\nu)}}
where in practice $\bar{\Ww}_{c_\phi,\epsilon}$ is approximated by minibatches and Sinkhorn, as mentioned above.

\paragraph{Putting everything together.} We can now describe efficiently our scheme and use Figure~\ref{fig-workflow} again for that purpose. In that figure, the generator (blue) and real data (green) parts are combined to compute a pairwise distance matrix $\tc$. This matrix, as in MMD-GAN's approach~\cite{li2015generative} is all we need. We do, however, significantly depart from a ``flat'' MMD approach in the red block of the figure, in which a finite number of Sinkhorn steps are used to approximate the Wasserstein distance. These Sinkhorn steps are used to evaluate (forward pass) and compute the gradient (backward pass) of that proxy as described in Algorithm~\ref{algo-sinkhorn}. Samples are repeatedly taken by taking push-forwards of samples of the initial measure in $\Zz$ to perform SGD as described in Algorithm~\ref{algo-sgd}. 


Note that the procedure $\texttt{AutoDiff}_\th$ corresponds to classical reverse mode automatic differentiation of $L$ steps of the Sinkhorn iteration, and has therefore naturally the same complexity as Sinkhorn, \emph{i.e.} $O(L\m\n)$ operations, with an extra storage cost required to run the backward iteration with no additional computational overhead.

The training procedure is the same as \cite{WassersteinGAN},\cite{MMDGAN} and consists in alterning $n_c$ optimisation steps to train the cost function $f_\phi$ and an optimisation step to train the generator $g_\th$. Following implementation advice from these papers,we clip the weights $\phi$ to ensure a bounded gradient in the maximization and use RMSProp as an optimizer.

\begin{algorithm}[H]
	    \caption{\label{algo-sgd}SGD with Auto-diff}
	 \begin{algorithmic}
	 	\Require $\th_0$, $\phi_0$, $(y_j)_{j=1}^\n$ (the real data), \m (batch size), $L$ (number of Sinkhorn iterations), $\epsilon$ (regularization parameter), $\alpha$ (learning rate)
		\Ensure $\th$, $\phi$ 
	\State $\th  \leftarrow \th_0$, $\phi  \leftarrow \phi_0$,
	\For{$k = 1,2,\dots$}
		\For {$t = 1,2,\dots,n_c$}
			\State Sample $(y_j)_{j=1}^{\m}$ from the observations.
			\State Sample $(z_i)_{i=1}^{\m} \overset{\text{i.i.d}}{\sim} \muz$, $(x_i)_{i=1}^{\m} \eqdef g_\theta(z_1^m)$
			\State $\texttt{grad}_\phi \leftarrow \texttt{AutoDiff}_\phi \Big( 2 \hat W_{\phi,\epsilon}^{(L)} (x_1^m,y_1^m)  $
			\State \hspace*{40pt} $ - \hat W_{\phi,\epsilon}^{(L)} (x_1^m,x_1^m) - \hat W_{\phi,\epsilon}^{(L)} (y_1^m,y_1^m) \Big) $
			\State $\phi \leftarrow \phi + \alpha \texttt{RMSProp}(\texttt{grad}_\phi)$.
			\State $\phi \leftarrow \texttt{clip}(\phi,-c,c)$ 
		\EndFor
		
		\State Sample $(y_j)_{j=1}^{\m}$ from the observations.
		\State Sample $(z_i)_{i=1}^{\m} \overset{\text{i.i.d}}{\sim} \muz$, $(x_i)_{i=1}^{\m} \eqdef g_\theta(z_1^m)$
		\State $\texttt{grad}_\th \leftarrow \texttt{AutoDiff}_\th \Big( 2 \hat W_{\phi,\epsilon}^{(L)} (x_1^m,y_1^m)  $
			\State \hspace*{40pt} $ - \hat W_{\phi,\epsilon}^{(L)} (x_1^m,x_1^m) - \hat W_{\phi,\epsilon}^{(L)} (y_1^m,y_1^m) \Big) $
		\State $\th \leftarrow \th - \alpha \texttt{RMSProp}(\texttt{grad}_\th)$.
	\EndFor
	\end{algorithmic}
	  \end{algorithm}

	  \begin{algorithm}[H]
	    \caption{\label{algo-sinkhorn}Sinkhorn loss $\hat W_{\phi,\epsilon}^{(L)} (x_1^m,y_1^m)$}
	 \begin{algorithmic}
	 	\Require $\th, (x_i)_{i=1}^\m, (y_j)_{j=1}^\m, \epsilon$ 
		\Ensure $w$ 
	\State $\foralls (i,j), \:  \tc_{i,j} \eqdef \norm{f_\phi(x_i)-f_\phi(y_j)}$
	\State $K_{i,j} = e^{-\frac{\tc_{i,j}}{\epsilon}}$
	\State $b  \leftarrow \ones_\n$,
	\For{$\ell = 1,2,\dots,L$}
		\State $a \leftarrow \frac{\ones_\n}{K b}$, 
			$b \leftarrow \frac{\ones_\n}{K^\top a}$
	\EndFor \\
	\Return $w = \dotp{(K\odot \tc) b}{a}$ (see \eqref{eq-return-value})
	\end{algorithmic}
	  \end{algorithm}
		

%% file: sections/applications.tex

\section{Applications}

We consider two popular problems in machine learning to illustrate the versatility of our method. The first one relies on fitting labeled data with uniform distribution supported on ellipses (note that this could be any parametric shape but ellipses here were a good fit). The second problem consists in tuning a neural network to generate images, first with a fixed cost (on MNIST dataset) and then with a parametric cost (on CIFAR10 dataset). \
In both cases, we used simple initializations (see details below) and the algorithm yielded similar results when rerun, meaning that the results displayed are representative of the performance of the algorithm and that the procedure is quite stable.

\subsection{Data Fitting with Ellipses.}

As mentioned earlier, a strength of the Wasserstein distance is its ability to fit a singular probability distribution to an empirical measure (data). That singular probability may be supported on a subset of the space on a lower dimensional manifold, or simply have a degenerate density that becomes null for some subsets of the original space. To illustrate this principle, we consider in what follows a simple 3D example that can easily be visualized.

We use the Iris dataset (3 classes, 50 observations each in 4 dimensions) projected in 3D using PCA. This defines the empirical measure $\nu$ in $\RR^3$. 
If we were to find a probability distribution $\mut$ bound to be itself an empirical measure of $K$ atoms (in that case parameter $\theta$ would contain exactly the locations of those $K$ points in addition to their weight), then minimizing the 2-Wasserstein distance of $\mu_\theta$ to $\nu$ would be \emph{strictly equivalent} to the $K$-means problem~\cite{NIPS2012_4651}. In that sense, quantization can be regarded as the most elementary example of Wasserstein loss minimization of degenerate families of probability distributions.

The model we consider is instead composed of $K$ ellipses with uniform density: Each ellipse is parametrized by a $3\times 3$ matrix $A_k$ (the square root of its covariance matrix) and a center $\alpha_k \in \RR^3$, so that $\th=(A_k,\al_k)_k$. Therefore, our results can't be directly compared to that of clustering algorithms, in the sense that we do automatically recover, within such ellipses, entire areas of interest (and not voronoi cells). We assume in this illustration that each ellipse has equal mass $1/K$.
To recover these ellipses through a push forward, we use a uniform ground density $\muz$ over $K$ centred unit balls, translated and dilated for each ellipse using the push-forward defined by $g_{\th}(z) = A_k z + \alpha_k$ if $z$ is in the $k$-th ball. 

\begin{table}
\centering
 \begin{tabular}{|c|c|c|c| } \hline
 MMD & $\epsilon = 1$ & $\epsilon = 0.1$ & $\epsilon = 0.001$ \\ \hline \hline
     $\begin{matrix} 36 & 0 & 0 \\ 0 & 39 & 13 \\
 0 & 11 & 42 \end{matrix}$
     &
     $\begin{matrix} 50 & 0 & 0 \\ 0 & 50 & 38 \\
0 & 36 & 47 \end{matrix}$
     & $\begin{matrix} 44 & 0 & 0 \\ 0 & 38 & 5 \\ 0 & 8 & 40 \end{matrix}$
     & $\begin{matrix}  33 & 0 & 0 \\  0 & 37 & 3 \\ 0 & 12 & 25\end{matrix}$
     \\\hline
    \end{tabular}
\caption{Evaluation of the fit after convergence of the algorithm : entry $(i,j)$ corresponds to the number of points from class $j$ that are inside ellipse $i$} \label{tab:clustering}
  \end{table}

  \begin{figure*}[ht]
  \centering
  \begin{subfigure}{0.25\textwidth}
  \includegraphics[width=\textwidth]{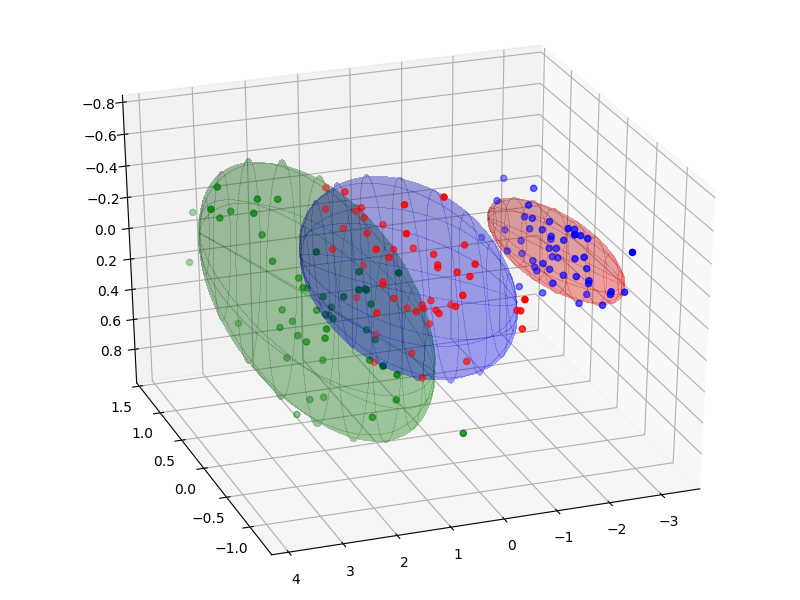} 
  \caption{MMD}
  \end{subfigure}%
  \begin{subfigure}{0.25\textwidth}
  \includegraphics[width=\textwidth]{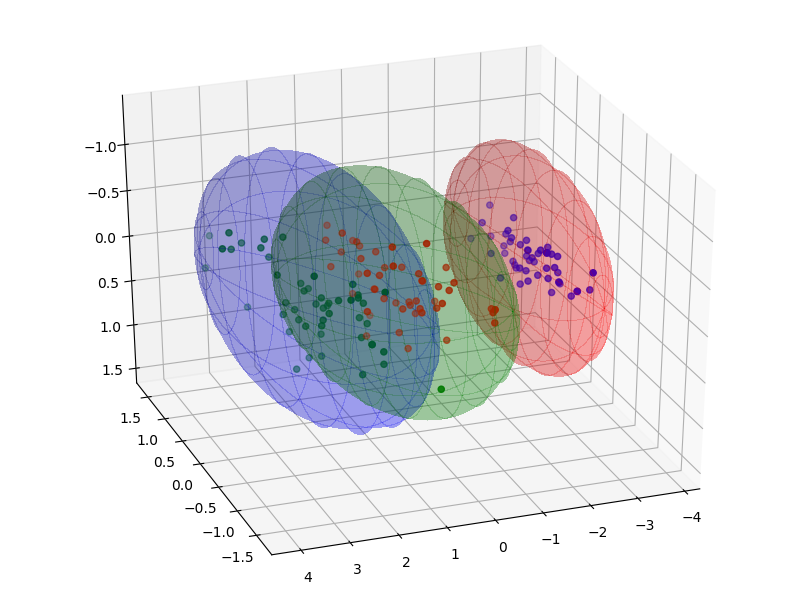} 
  \caption{$\epsilon = 1$ }
  \end{subfigure}
  \begin{subfigure}{0.25\textwidth}
  \includegraphics[width=\textwidth]{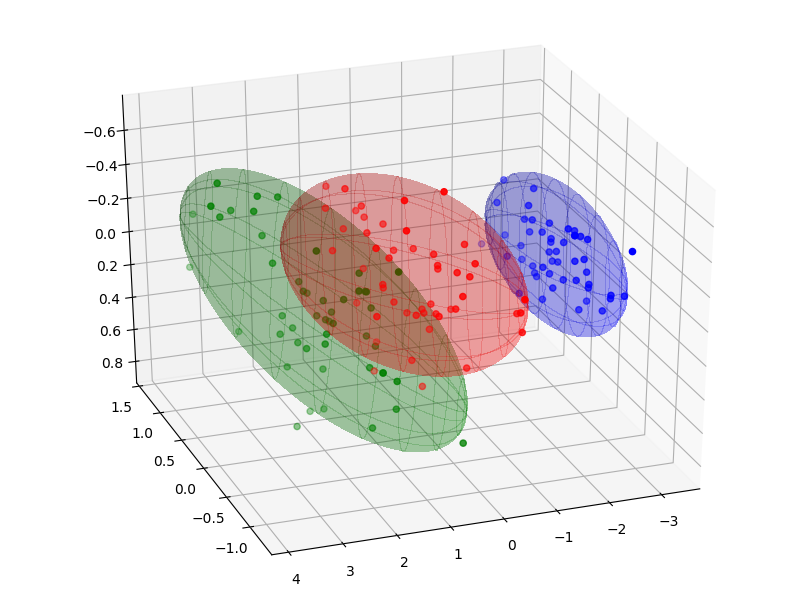} 
  \caption{$\epsilon = 0.1$}
  \end{subfigure}%
  \begin{subfigure}{0.25\textwidth}
  \includegraphics[width=\textwidth]{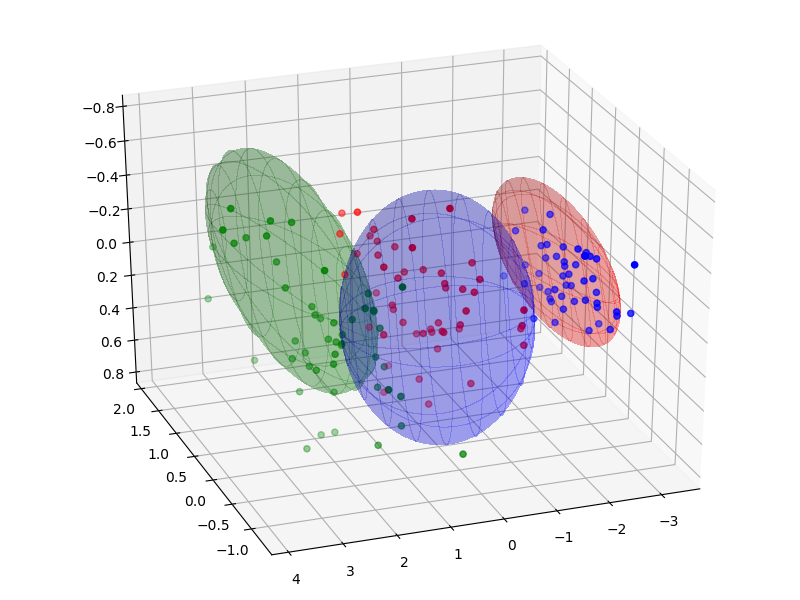} 
  \caption{$\epsilon = 0.01$}
  \end{subfigure}%
  \caption{Ellipses after convergence of the stochastic gradient descent with $L=20$, $m=200$
  \label{fig:kballs}}
  \end{figure*}
\paragraph{Numerical Illustration.} The fit is obtained using the cost $c(x,y)=\norm{x-y}^2$, the ellipse matrices $(A_k)_k$ are all initialized with the identity matrix (which corresponds to the unit ball) and centers $(\al_k)_k$ are initialized with the $K$-means algorithm. We fixed a maximal buget of Sinkhorn iterations $L = 20$ to be competitive with MMD time-wise, with a minibatch size $m=200$ for both algorithms. Figure~\ref{fig:kballs} displays the results of our method for different values of $\epsilon$ and for MMD with a gaussian kernel (with manually tuned bandwith ). The influence of the regularization parameter $\epsilon$ is crucial: too much regularization (large $\epsilon$,(b)) leads to a loose fit of the data but not regularizing enough leads to very slow convergence of the Sinkhorn algorithm and also yeilds poor performance (d) or requires more cpu time if we increase the total iteration budget. 
Since the Iris data is labeled, we can asses the fit of the model by checking the class repartition in each ellipse, as summarized in table \ref{tab:clustering}. Each entry $(i,j)$ corresponds to the number of points from class $j$ that are inside ellipse $i$ (recall there are 50 points per class). The performance difference between MMD and Sinkhorn here is not obvious, once the bandwidth parameter of the kernel is carefully tuned, but we found out that this parameter was more sensitive than $\epsilon$, as the range of values that yield acceptable results are smaller.


\subsection{Tuning a Generative Neural Network}

Image generating models such as GAN~\cite{GAN} or VAE~\cite{VAE} have become popular in recent years. The goal is to train a neural network $g_\th$ which generates images $g_\th(z)$ that resemble a certain data set $(y_j)_j$, given a random input $z$ in a latent space $\Zz$. Both methods require a second network for the training of the generative network (an adversial network in the case of GANs, an encoding network in the case of VAEs). Depending on the complexity of the data, our method can rely on the generative network alone by directly comparing its output with the data in Wasserstein distance.

\paragraph{With a fixed cost $c$} This section fits a generative model where the pushforward $g_\th$ is a multilayer perceptron. We begin with experiments on the MNIST dataset, which is a standard benchmark for this type of networks. Since the dataset is relatively simple, learning the cost is superfluous here and we use the ground cost $c(x,y)=\norm{x-y}^2$, which is sufficient for these low resolution images and also the baseline in~\cite{VAE}. We use as $g_\th$ a multilayer perceptron with a 2D latent space $\Zz=\RR^2$. It is composed of 2 fully connected layers: one hidden layer of 500 units and an output layer which is the same size as the data ($\Xx = \RR^{28 \times 28}$). The parameters $\theta$ are thus the weights of both layers, and are initialized with the Xavier method~\cite{Xavier}. We choose $\zeta$ to be a uniform distribution over the unit square $[0,1]^2$. Learning is performed in mini-batches over the MNIST dataset, with the Adam optimizer~\cite{kingma2014adam}.

\begin{figure*}
\centering
\begin{subfigure}{0.28\linewidth}
\includegraphics[width=\linewidth]{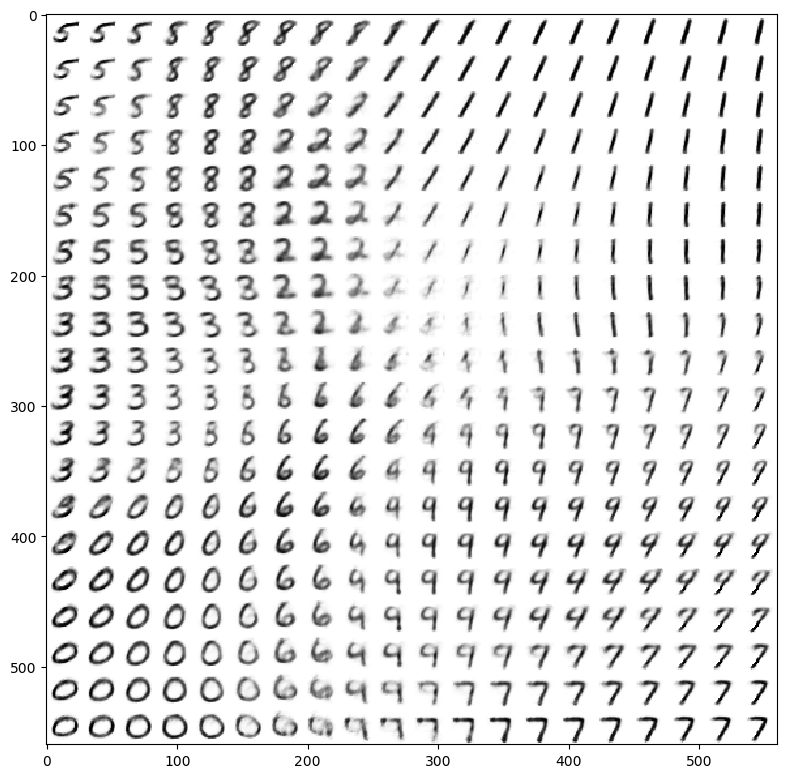} 
\caption*{\hspace*{5pt}(a) $\epsilon = 1$ \\ \hspace*{5pt}$m = 200 , L = 10$}
\end{subfigure}%
\quad
\begin{subfigure}{0.28\linewidth}
\includegraphics[width=\linewidth]{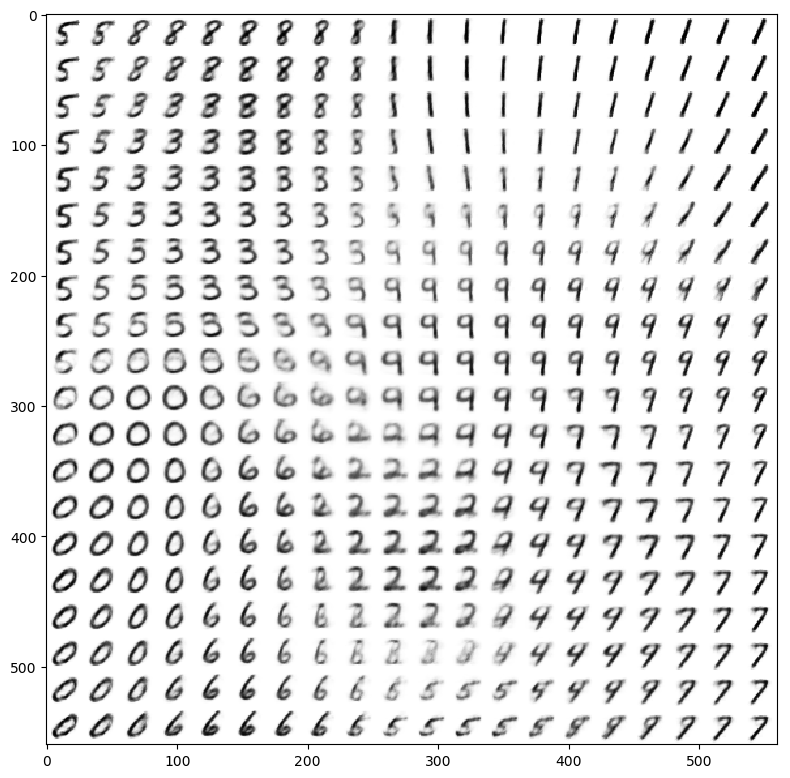} 
\caption*{\hspace*{5pt}(b) $\epsilon = 10^{-1}$ \\ \hspace*{5pt}$m = 200 , L = 100$}
\end{subfigure}
\quad
\begin{subfigure}{0.28\linewidth}
\includegraphics[width=\linewidth]{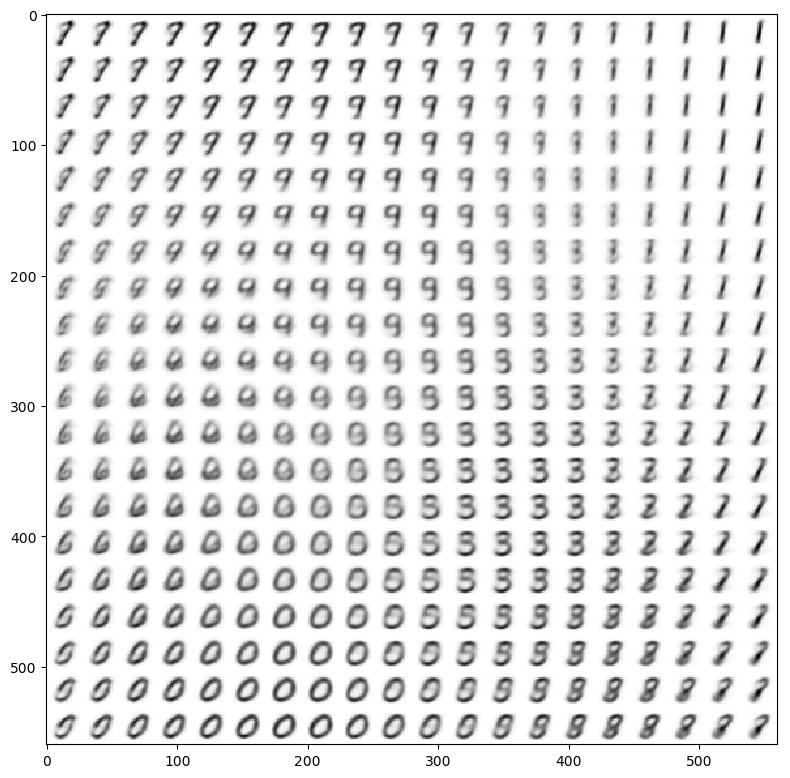} 
\caption*{\hspace*{5pt}(c) $\epsilon = 10^{-1}$\\ \hspace*{5pt}$m = 10, L = 300$}
\end{subfigure}%
\caption{Influence of the hyperparameters on the manifold of generated digits.
\label{fig:MNIST}}
\end{figure*}

Figure~\ref{fig:MNIST} displays the manifold of images $g_\th(z)$ generated by the optimized network (i.e. for equi-spaced $z \in [0,1]^2$) after the learning procedure for different values of the hyperparameters $(\epsilon,m,L)$. This shows that the regularization parameter $\epsilon$ can be chosen quite large, which in turn leads to a fast convergence of Sinkhorn iterations. Indeed, using $\epsilon = 1$ with only $L=10$ Sinkhorn iterations (image~(a)) yields a result similar to using $\epsilon = 0.1$ with $L=100$ iterations (image (b)). Regarding the size $m$ of the mini-batches, a too small $m$ value (e.g. $m=10$) leads to poor results, and we observe that $m=200$ is sufficient to learn accurately the manifold. 

\begin{figure*}[ht]
\centering
\begin{subfigure}{0.3\linewidth}
\includegraphics[width=\linewidth]{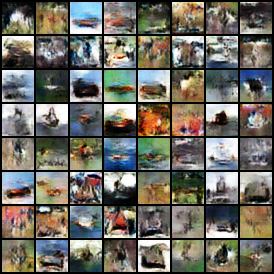} 
\caption{MMD}
\end{subfigure}%
\hspace*{7pt}
\begin{subfigure}{0.3\linewidth}
\includegraphics[width=\linewidth]{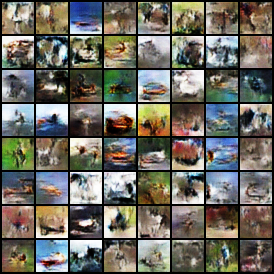} 
\caption{$\epsilon = 1000$}
\end{subfigure}%
\hspace*{7pt}
\begin{subfigure}{0.3\linewidth}
\includegraphics[width=\linewidth]{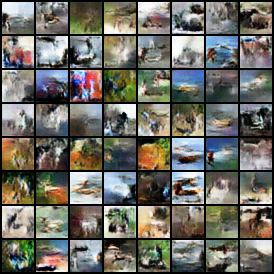} 
\caption{$\epsilon = 10$}
\end{subfigure}%
\caption{Samples from the generator trained on CIFAR 10 for MMD and Sinkhorn loss (coming from the same samples in the latent space)}
\label{fig:cifar10}
\end{figure*}

\paragraph{Learning the cost} With higher-resolution datasets, such as classical benchmarks CIFAR10 or CelebA, using the $\ell^2$ metric between images yields very poor results. It tends to generate images which are basically a blur of similar images. The alternative, already outlined in Algortithm 1 relies on learning another network wich encodes meaningful feature vectors for the images, between which can take the euclidean distance.

We compare our loss with different values for the regularization parameter $\epsilon$ to the results obtained with an MMD loss with a gaussian kernel. The experimental setting is the same as in~\cite{MMDGAN} and we used the same parameters to carry out a fair comparison.

\begin{table}
\centering
 \begin{tabular}{c c c c  }
        MMD  & $\epsilon = 1000$ & $\epsilon = 100$ & $\epsilon = 10$   \\  \hline 
     $4.04 \pm 0.07$ & $4.14 \pm 0.06$  &  $3.09 \pm 0.036$    &  $3.11 \pm 0.031$     
    \end{tabular}
\caption{Inception Scores} \label{tab:inception}
  \end{table}

Table~\ref{tab:inception} summarizes the inception scores on CIFAR10 for MMD and Sinkhorn loss with varying $\epsilon$. Generative models are very hard to evaluate and there is no consensus on which metric should be used to assess their quality. We choose the inception score introduced in~\cite{inception} as it is well spread, and also the reference in~\cite{MMD-GAN} agains which we compare our losses. The scores are evalutated on 20000 random images.  Figure~\ref{fig:cifar10} displays a few of the associated samples (generated with the same seed). Although there is no striking difference in visual quality, the model with a Sinkhorn loss and a large regularization is the one with the best score. The poor scores of models which have a loss closer to the true OT loss can be explained by two main factors : (i)~the number of iterations required for the convergence of Sinkhorn with such $\epsilon$ might exceed the total iteration budget that we give the algorithm to compute the loss (to ensure reasonable training time of the model), (ii)~it reflects the fact that sample complexity worsens when we get closer to OT metrics, and increasing the batch size might be beneficial in that case.

%% file: sections/conclusion.tex

\section*{Conclusion}

In this paper, we presented a new computational toolbox to train large scale generative models with the Sinkhorn divergence.
Thanks to the combination of entropic smoothing and automatic differentiation, it makes optimal transport applicable in arbitrary complex generative model setups.
Besides, we proved that this divergence interpolates between classical OT and MMD losses, benefiting from advantages of both frameworks. 
Future work should focus on theoretical properties of the Sinkhorn divergence, in particular sample complexity and positivity.

%% file: sections/numerics-div.tex

\section{Numerical Exploration of the Sinkhorn Divergence}

\subsection{Sample Complexity}

To better grasp the statistical tradeoff offered by the entropic regularization, we study numerically the so-called sample complexity of these divergence.
We consider 
\eq{
	\hat \mu_N = \frac{1}{N}\sum_{i=1}^N \de_{x_i}
	\qandq
	\hat \nu_N = \frac{1}{N}\sum_{i=1}^N \de_{x_i}
}
which are random measures, where the $(x_i)_i$ and $(y_i)_i$ are ponts independently drawn from the same distribution $\xi$.
In the numerical experiments, $\xi$ is the uniform distribution on $[0,1]^d$ where $d \in \NN^*$ is the ambient dimension. 

We recall that 
\eq{
	\bar\Ww_{c,\epsilon}(\mu,\nu) \eqdef 2\Ww_{c,\epsilon}(\mu,\nu)-\Ww_{c,\epsilon}(\mu,\mu)-\Ww_{c,\epsilon}(\nu,\nu)
}
\eq{
	\qwhereq 
	\Ww_{c,\epsilon}(\mu,\nu) \eqdef \int c(x,y) \d\ga_{\epsilon}
}
where $\ga_{\epsilon}$ is the unique solution of the entropy-regularization optimal transport problem between $\mu$ and $\nu$.
In the following, we consider $c(x,y)=\norm{x-y}^p$ for $p=3/2$ for $(x,y) \in (\RR^{d})^2$.

\begin{figure*}
\centering
\begin{tabular}{c@{}c@{}c}
\includegraphics[width=.32\linewidth]{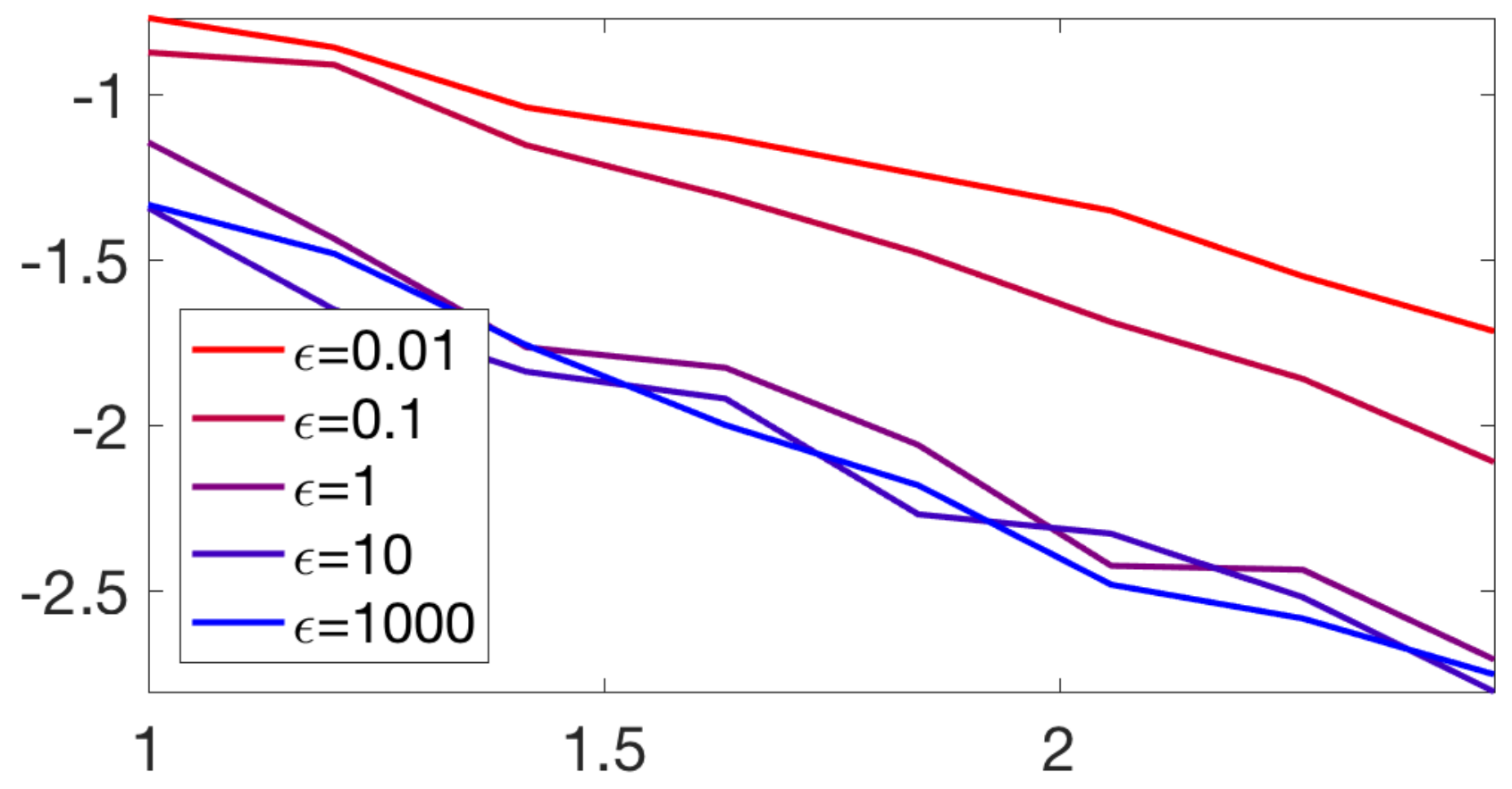}&
\includegraphics[width=.32\linewidth]{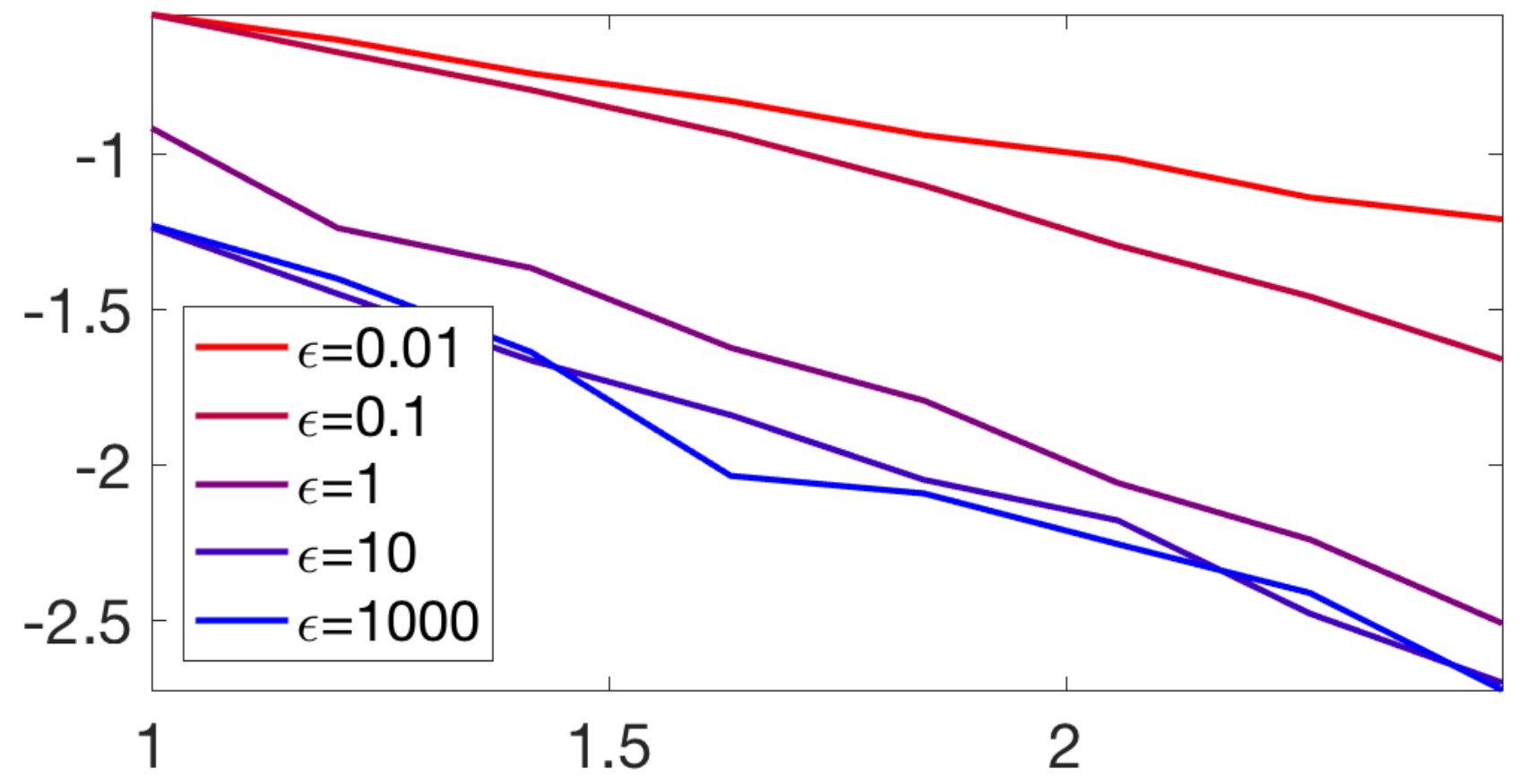}&
\includegraphics[width=.32\linewidth]{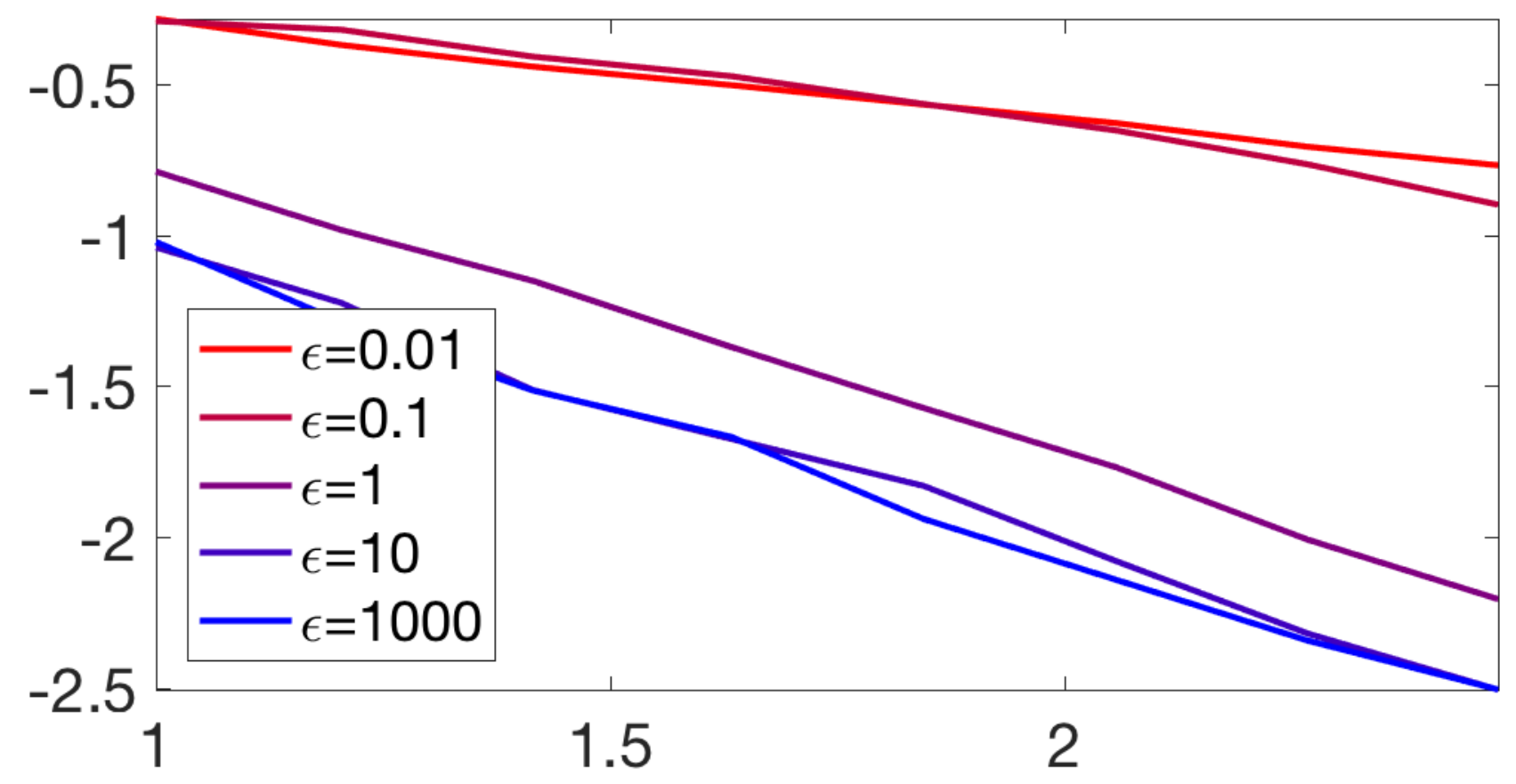}\\
$d=2$& $d=3$ & $d=5$\\
\end{tabular}%
\caption{ Influence of the regularization $\epsilon$on the sample complexity rate. The plot displays $\log_{10}(R_{\epsilon,d}(N))$ as a function of $\log(N)$. 
\label{fig:influ-eps}}
\end{figure*}

One has the convergence
\begin{align*}
	\Ww_{c,\epsilon}(\mu,\nu) &\overset{\epsilon\rightarrow 0}{\longrightarrow} 2 W_p(\mu,\nu)^p \\
	\qandq
	\Ww_{c,\epsilon}(\mu,\nu) &\overset{\epsilon\rightarrow +\infty}{\longrightarrow} \norm{\mu-\nu}_{\text{ED}(p)}^2
\end{align*}
where $W_p$ is the Wasserstein-$p$ distance while $\norm{\xi}_{\text{ED}(p)}^2  = \int -\norm{x-y}^p \d\xi(x)\d\xi(y)$ is the Energy Distance, which is a special case of MMD norm for $0<p<2$.

The goal is to study numerically the decay rate toward zero of 
\eq{
	R_{\epsilon,d}(N) \eqdef \EE(\bar \Ww_{c,\epsilon}(\hat\mu_N,\hat\nu_N))
}
and also analyze the standard deviation
\eq{
	S_{\epsilon,d}^2(N) \eqdef \EE( |\bar \Ww_{c,\epsilon}(\hat\mu_N,\hat\nu_N)-R_{\epsilon,d}(N)|^2 ).
}
In these formula, the expectation $\EE$ with respect to random draws of $(x_i)_i$ and $(y_i)_i$ is estimated numerically by averaging over $10^3$ drawings.
For optimal transport, i.e. $\epsilon=0$, it is well-known (we refer to the references given in the paper) that $R_{0,d}(N) = O(\frac{1}{N^{p/d}})$, while for MMD norm, i.e. $\epsilon=+\infty$, one has $R_{+\infty,d}(N) = O(\frac{1}{N})$.

\begin{figure*}
\centering
\begin{tabular}{@{}c@{}c@{}c@{}}
\includegraphics[width=.32\linewidth]{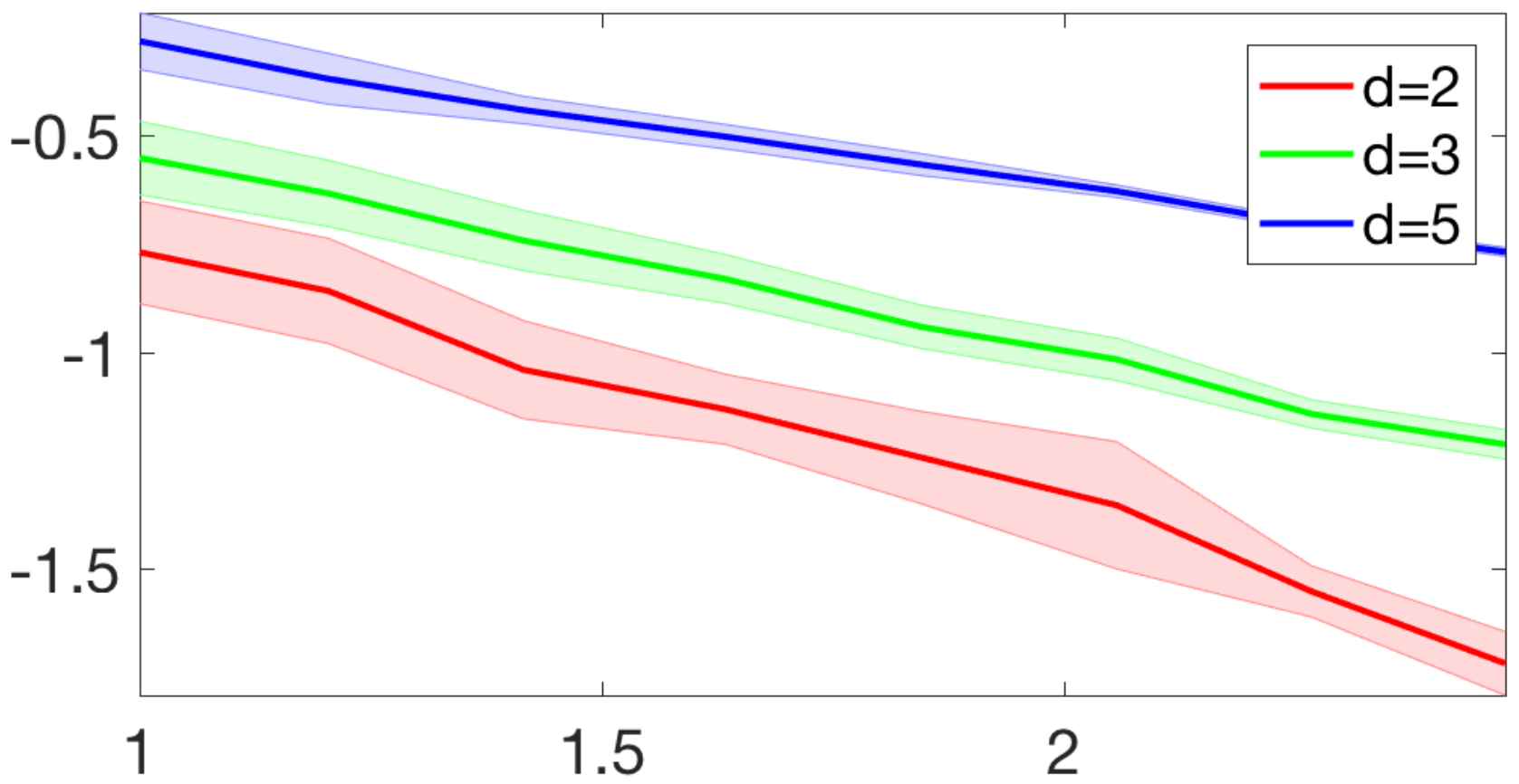}&
\includegraphics[width=.32\linewidth]{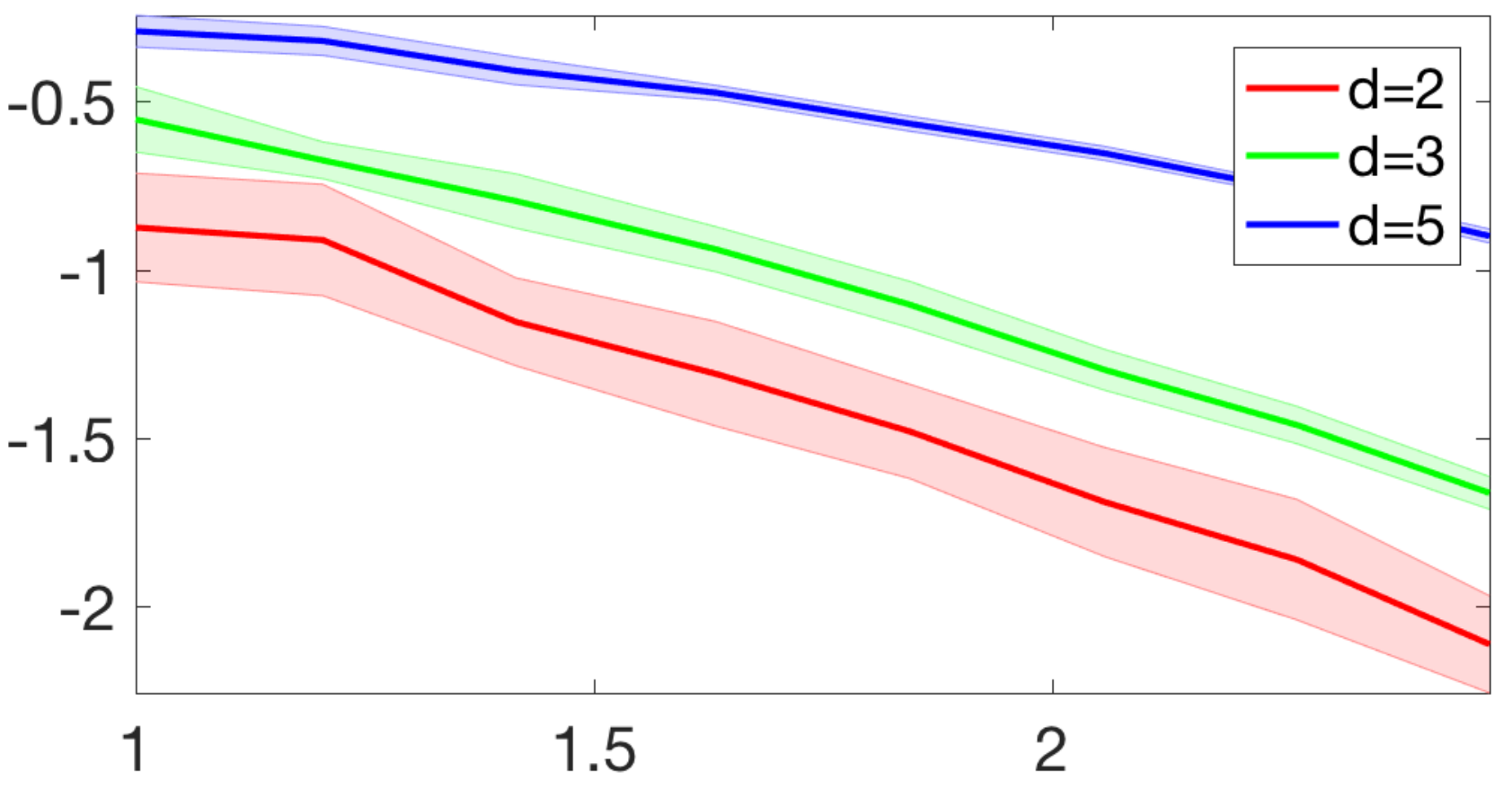}&
\includegraphics[width=.32\linewidth]{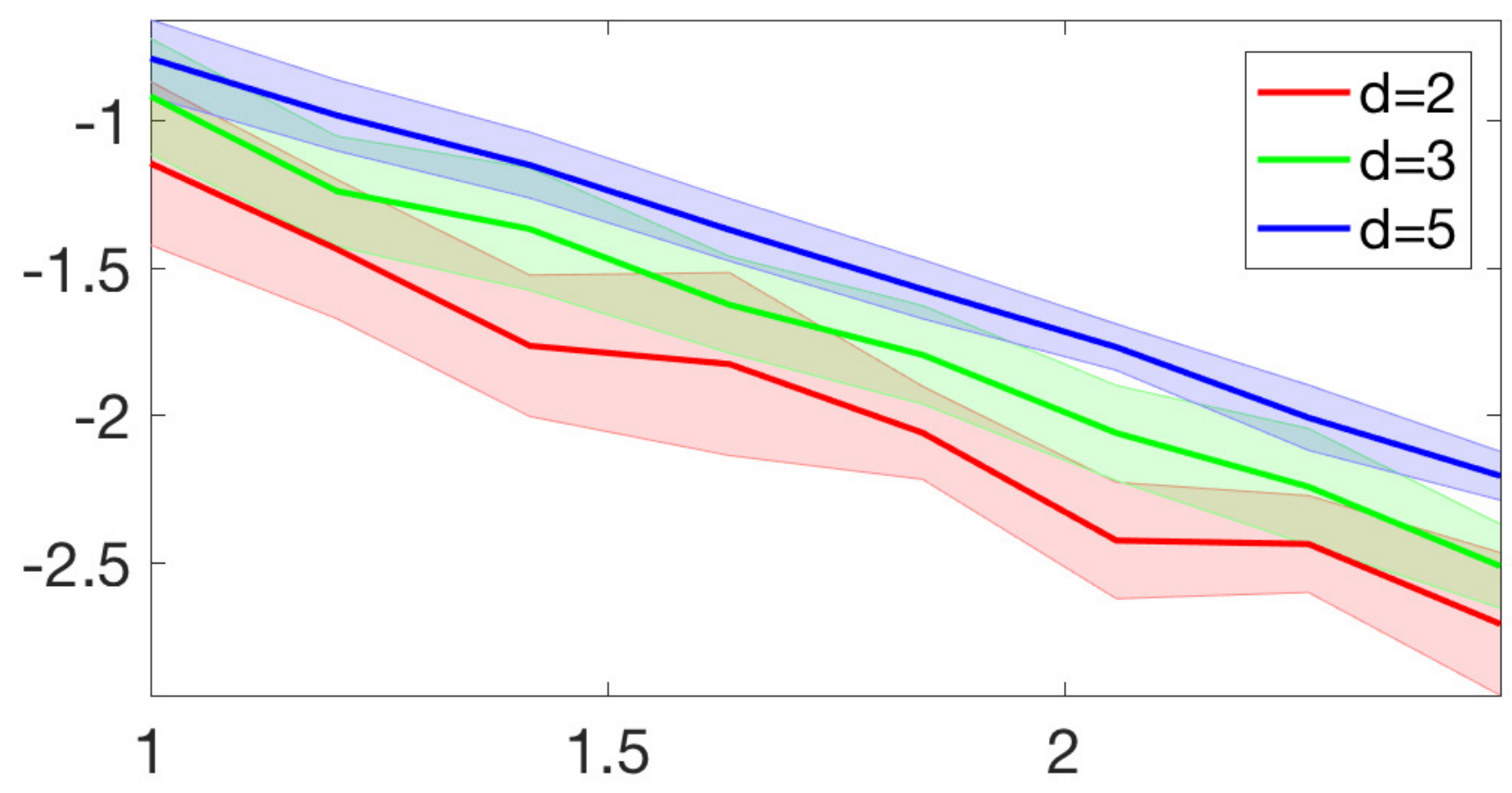}\\
$\epsilon=0.01$& $\epsilon=.1$ & $\epsilon=1$ 
\end{tabular}%
\caption{ Influence of the dimension $d$  on the sample complexity rate for difference $d$. The plot displays $\log_{10}(R_{\epsilon,d}(N))$ as a function of $\log(N)$. 
The shaded bar display the confidence interval at $\pm S_{\epsilon,d}(N)$.
\label{fig:influ-d}}
\end{figure*}

Figure~\ref{fig:influ-d} (resp.~\ref{fig:influ-eps}) display in log-log plot the decay of $R_{\epsilon,d}(N)$ with $N$, and allows to compare on a single plot the influence of $d$ (resp. $\epsilon$) for a fixed $\epsilon$ (resp. $d$) on each plot.

From these experiments, one can conclude on this distribution $\xi$ that:
\begin{itemize}
	\item $\Ww_{c,\epsilon}(\mu,\nu) \geq 0$ (more on this in the following section).
	\item $R_{\epsilon,d}(N)$ as a polynomial decay of the form $1/N^{\kappa_{\epsilon,d}}$.
	\item One recovers the known rates $\kappa_{0,d}=p/d$ (here for $p=3/2$) and $\kappa_{\infty,d}=1$.	
	\item Small values of $\epsilon < 1$ have rates $\kappa_{\epsilon,d}$ close to the rate of OT $\kappa_{0,d}$.
	\item Large values of $\epsilon > 1$ have rates $\kappa_{\epsilon,d}$ matching almost exactly the rate of MMD $\kappa_{+\infty,d}=1$.
	\item The variance $S_{\epsilon,d}^2(N)$ is significantly smaller for small values of $\epsilon$ (i.e. close to OT). 
\end{itemize}
Note that similar conclusion are obtained when testing on other distributions $\xi$ (e.g. a Gaussian).

\subsection{Positivity}

For $\epsilon \in \{0,+\infty\}$, both OT and MMD are distances, so that $\bar\Ww_{\epsilon,c}(\mu,\nu)=0$ if and only if $\mu=\nu$.
It not known whether this property is true for $0 < \epsilon < +\infty$, and this seems a very difficult problem to tackle. 
We investigate numerically this question by looking at small modification of a discrete input measure $\mu = \frac{1}{\sum_i a_i}\sum_{i=1}^N a_i \de_{x_i}$ where the $x_i$ are i.i.d. points drawn in $[0,1]^2$ and $(a_i)_i$ are i.i.d. number drawn uniformly in $[1/2,1]$, and perform a small modification
\eq{
	\mu_t \eqdef \frac{1}{\sum_i a_{i,t}}\sum_{i=1}^N a_i \de_{x_{i,t}}
	\qwhereq
	\choice{
		a_{i,t}=a_{i,t}+t b_i, \\
		x_{i,t}=x_i+t z_i, \\
	}
}
where $(b_i)_i \subset \RR$ are i.d.d. Gaussian distributed $\Nn(0,1)$ and
where $(z_i)_i \subset \RR^2$ are i.d.d. Gaussian distributed $\Nn(0,\Id_2)$.

\begin{figure*}
\centering
\begin{tabular}{c@{\hspace{5mm}}c}
\includegraphics[width=.32\linewidth]{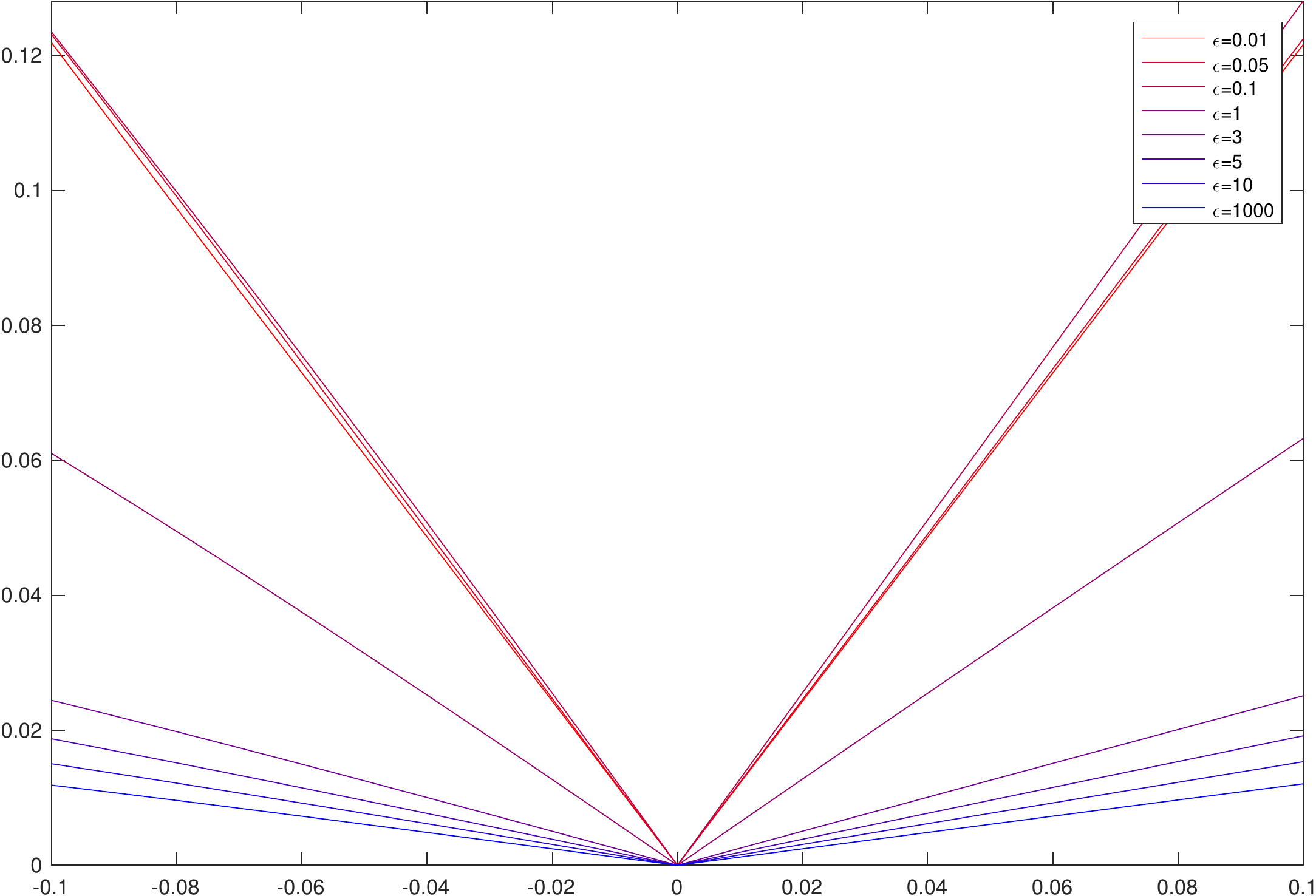}&
\includegraphics[width=.32\linewidth]{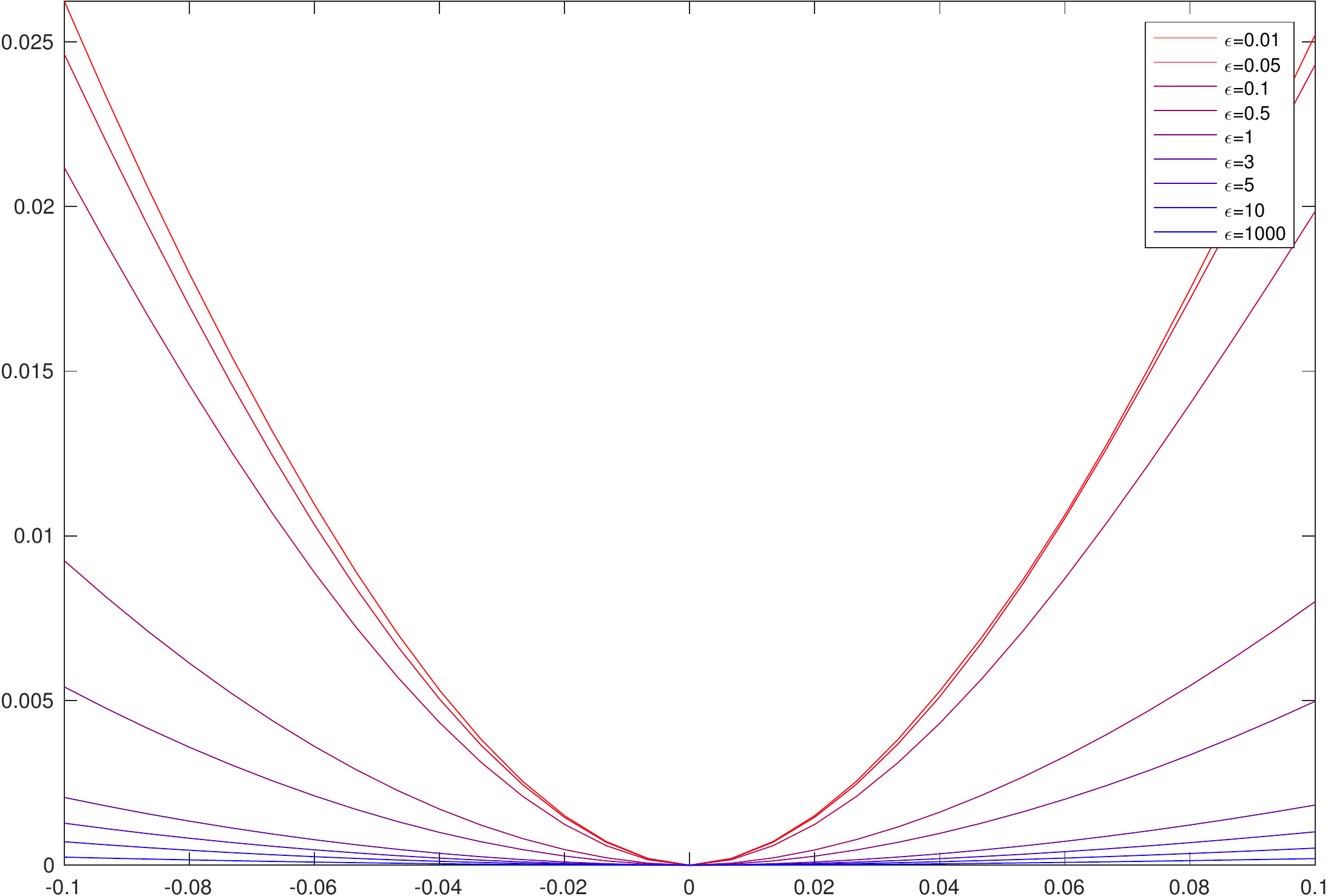}\\
$p=1$, $N=10$ & $p=1.8$, $N=100$
\end{tabular}%
\caption{ Test of the positivity of $\bar\Ww_{\epsilon,c}(\mu,\mu_t)$ as a function of the perturbation parameter~$t$. 
\label{fig:positivity}}
\end{figure*}

Figure~\eqref{fig:positivity} shows, on a single realization of $(a_i,x_i,b_i,z_i)$, that $\bar\Ww_{\epsilon,c}(\mu,\mu_t)>0$ for $t \neq 0$. Testing for $10^4$ other realizations gives the same results, showing that experimentally  $\bar\Ww_{\epsilon,c}$ is locally strictly positive for discrete measures. 

